# Sentiment Analysis

## A Literature Survey

*by*

Subhabrata Mukherjee

Roll No: 10305061

Supervisor: Dr. Pushpak Bhattacharyya

June 29, 2012

Indian Institute of Technology, Bombay

Department of Computer Science and Engineering

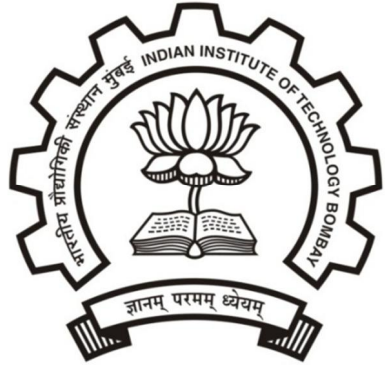

# Abstract

Our day-to-day life has always been influenced by what people think. Ideas and opinions of others have always affected our own opinions. The explosion of Web 2.0 has led to increased activity in Podcasting, Blogging, Tagging, Contributing to RSS, Social Bookmarking, and Social Networking. As a result there has been an eruption of interest in people to mine these vast resources of data for opinions. Sentiment Analysis or Opinion Mining is the computational treatment of opinions, sentiments and subjectivity of text. In this report, we take a look at the various challenges and applications of Sentiment Analysis. We will discuss in details various approaches to perform a computational treatment of sentiments and opinions. Various supervised or data-driven techniques to SA like Naïve Byes, Maximum Entropy, SVM, and Voted Perceptrons will be discussed and their strengths and drawbacks will be touched upon. We will also see a new dimension of analyzing sentiments by Cognitive Psychology mainly through the work of Janyce Wiebe, where we will see ways to detect subjectivity, perspective in narrative and understanding the discourse structure. We will also study some specific topics in Sentiment Analysis and the contemporary works in those areas.

# TABLE OF CONTENTS

# Table of Figures

# 1. INTRODUCTION

## 1.1 WHAT IS SENTIMENT ANALYSIS?

Sentiment Analysis is a *Natural Language Processing* and *Information Extraction* task that aims to obtain writer's feelings expressed in positive or negative comments, questions and requests, by analyzing a large numbers of documents. Generally speaking, sentiment analysis aims to determine the attitude of a speaker or a writer with respect to some topic or the overall tonality of a document. In recent years, the exponential increase in the Internet usage and exchange of public opinion is the driving force behind Sentiment Analysis today. The Web is a huge repository of structured and unstructured data. The analysis of this data to extract latent public opinion and sentiment is a challenging task.

Liu *et al.* (2009) defines a sentiment or opinion as a quintuple-

"<$o_j, f_{jk}, so_{ijkl}, h_i, t_l$ >, *where $o_j$ is a target object, $f_{jk}$ is a feature of the object $o_j$, $so_{ijkl}$ is the sentiment value of the opinion of the opinion holder $h_i$ on feature $f_{jk}$ of object $o_j$ at time $t_l$, $so_{ijkl}$ is +ve,-ve, or neutral, or a more granular rating, $h_i$ is an opinion holder, $t_l$ is the time when the opinion is expressed.*"

The analysis of sentiments may be document based where the sentiment in the entire document is summarized as positive, negative or objective. It can be sentence based where individual sentences, bearing sentiments, in the text are classified. SA can be phrase based where the phrases in a sentence are classified according to polarity.

Sentiment Analysis identifies the phrases in a text that bears some sentiment. The author may speak about some *objective facts* or *subjective opinions*. It is necessary to distinguish between the two. SA finds the subject towards whom the sentiment is directed. A text may contain many entities but it is necessary to find the entity towards which the sentiment is directed. It identifies the polarity and degree of the sentiment. Sentiments are classified as *objective* (facts), *positive* (denotes a state of happiness, bliss or satisfaction on part of the writer) or *negative* (denotes a state of sorrow, dejection or disappointment on part of the writer). The sentiments can further be given a score based on their *degree* of positivity, negativity or objectivity.

## 1.2 APPLICATIONS OF SENTIMENT ANALYSIS

Word of mouth (WOM) is the process of conveying information from person to person and plays a major role in customer buying decisions. In commercial situations, WOM involves consumers sharing attitudes, opinions, or reactions about businesses, products, or services with other people. WOM communication functions based on social networking and trust. People rely on families, friends, and others in their social network. Research also indicates that people appear to trust seemingly disinterested opinions from people outside their immediate social network, such as online reviews. This is where Sentiment Analysis comes into play. Growing availability of opinion rich resources like online review sites, blogs, social networking sites have made this "decision-making process" easier for us. With explosion of Web 2.0 platforms consumers have a soapbox of unprecedented reach and power by which they can share opinions. Major companies have realized these consumer voices affect shaping voices of other consumers.

Sentiment Analysis thus finds its use in *Consumer Market* for Product reviews, *Marketing* for knowing consumer attitudes and trends, *Social Media* for finding general opinion about recent hot topics in town, *Movie* to find whether a recently released movie is a hit.

Pang-Lee *et al.* (2002) broadly classifies the applications into the following categories.

a. Applications to Review-Related Websites

   Movie Reviews, Product Reviews *etc.*

b. Applications as a Sub-Component Technology

   Detecting antagonistic, heated language in mails, spam detection, context sensitive information detection *etc.*

c. Applications in Business and Government Intelligence

   Knowing Consumer attitudes and trends

d. Applications across Different Domains

   Knowing public opinions for political leaders or their notions about rules and regulations in place *etc.*

## 1.3  CHALLENGES FOR SENTIMENT ANALYSIS

Sentiment Analysis approaches aim to extract *positive* and *negative* sentiment bearing words from a text and classify the text as *positive, negative* or else *objective* if it cannot find any sentiment bearing words. In this respect, it can be thought of as a text categorization task. In text classification there are many classes corresponding to different topics whereas in Sentiment Analysis we have only 3 broad classes. Thus it seems Sentiment Analysis is easier than text classification which is not quite the case. The general challenges can be summarized as:

### 1.3.1  Implicit Sentiment and Sarcasm

A sentence may have an implicit sentiment even without the presence of any sentiment bearing words. Consider the following examples.

> *How can anyone sit through this movie?*

> *One should question the stability of mind of the writer who wrote this book.*

Both the above sentences do not explicitly carry any negative sentiment bearing words although both are negative sentences. Thus *identifying semantics* is more important in SA than *syntax detection.*

### 1.3.2  Domain Dependency

There are many words whose polarity changes from domain to domain. Consider the following examples.

> *The story was unpredictable.*

> *The steering of the car is unpredictable.*

> *Go read the book.*

In the first example, the sentiment conveyed is positive whereas the sentiment conveyed in the second is negative. The third example has a positive sentiment in the book domain but a negative sentiment in the movie domain (where the director is being asked to go and read the book).

### 1.3.3 Thwarted Expectations

Sometimes the author deliberately sets up context only to refute it at the end. Consider the following example:

*This film should be brilliant. It sounds like a great plot, the actors are first grade, and the supporting cast is good as well, and Stallone is attempting to deliver a good performance. However, it can't hold up.*

Inspite of the presence of words that are positive in orientation the overall sentiment is negative because of the crucial last sentence, whereas in traditional text classification this would have been classified as positive as *term frequency* is more important there than *term presence*.

### 1.3.4 Pragmatics

It is important to detect the pragmatics of user opinion which may change the sentiment thoroughly. Consider the following examples:

*I just finished watching Barca DESTROY Ac Milan*

*That final completely destroyed me.*

Capitalization can be used with subtlety to denote sentiment. The first example denotes a positive sentiment whereas the second denotes a negative sentiment. There are many other ways of expressing pragmatism.

### 1.3.5 World Knowledge

Often world knowledge needs to be incorporated in the system for detecting sentiments. Consider the following examples:

*He is a Frankenstein.*

*Just finished Doctor Zhivago for the first time and all I can say is Russia sucks.*

The first sentence depicts a negative sentiment whereas the second one depicts a positive sentiment. But one has to know about *Frankenstein and Doctor Zhivago* to find out the sentiment.

### 1.3.6 Subjectivity Detection

This is to differentiate between opinionated and non-opinionated text. This is used to enhance the performance of the system by including a subjectivity detection module to filter out objective facts. But this is often difficult to do. Consider the following examples:

*I hate love stories.*

*I do not like the movie "I hate stories".*

The first example presents an objective fact whereas the second example depicts the opinion about a particular movie.

### 1.3.7 Entity Identification

A text or sentence may have multiple entities. It is extremely important to find out the entity towards which the opinion is directed. Consider the following examples.

*Samsung is better than Nokia*

*Ram defeated Hari in football.*

The examples are positive for Samsung and Ram respectively but negative for Nokia and Hari.

### 1.3.8 Negation

Handling negation is a challenging task in SA. Negation can be expressed in subtle ways even without the explicit use of any negative word. A method often followed in handling negation explicitly in sentences like "*I do not like the movie*", is to reverse the polarity of all the words appearing after the negation operator (like *not)*. But this does not work for "*I do not like the acting but I like the direction*". So we need to consider the *scope* of negation as well, which extends only till *but* here. So the thing that can be done is to change polarity of all words appearing after a negation word till another negation word appears. But still there can be problems. For example, in the sentence "*Not only did I like the acting, but also the direction*", the polarity is *not reversed* after "not" due to the presence of "*only*". So this type of combinations of "not" with other words like "only" has to be kept in mind while designing the algorithm.

## 1.4 FEATURES FOR SENTIMENT ANALYSIS

Feature engineering is an extremely basic and essential task for Sentiment Analysis. Converting a piece of text to a feature vector is the basic step in any data driven approach to SA. In the following section we will see some commonly used features used in Sentiment Analysis and their critiques.

**Term Presence vs. Term Frequency**

Term frequency has always been considered essential in traditional Information Retrieval and Text Classification tasks. But Pang-Lee *et al.* (2002) found that *term presence* is more important to Sentiment analysis than *term frequency*. That is, binary-valued feature vectors in which the entries merely indicate whether a term occurs (value 1) or not (value 0). This is not counter-intuitive as in the numerous examples we saw before that the presence of even a single string sentiment bearing words can reverse the polarity of the entire sentence. It has also been seen that the occurrence of rare words contain more information than frequently occurring words, a phenomenon called *Hapax Legomena*.

**Term Position**

Words appearing in certain positions in the text carry more sentiment or weightage than words appearing elsewhere. This is similar to IR where words appearing in topic Titles, Subtitles or Abstracts *etc* are given more weightage than those appearing in the body. In the example given in Section 1.3.c, although the text contains positive words throughout, the presence of a negative sentiment at the end sentence plays the deciding role in determining the sentiment. Thus generally words appearing in the 1$^{st}$ few sentences and last few sentences in a text are given more weightage than those appearing elsewhere.

**N-gram Features**

N-grams are capable of capturing context to some extent and are widely used in Natural Language Processing tasks. Whether higher order n-grams are useful is a matter of debate. Pang *et al.* (2002) reported that unigrams outperform bigrams when classifying movie reviews by sentiment polarity, but Dave *et al.* (2003) found that in some settings, bigrams and trigrams perform better.

**Subsequence Kernels**

Most of the works on Sentiment Analysis use word or sentence level model, the results of which are averaged across all words/sentences/n-grams in order to produce a single model output for each review. Bikel *et al.* (2007) use *subsequences.* The intuition is that the feature space implicitly captured by subsequence kernels is sufficiently rich to obviate the need for explicit knowledge engineering or modeling of word- or sentence-level sentiment.

Word sequence kernels of order n are a weighted sum over all possible word sequences of length n that occur in both of the strings being compared.

Mathematically, the word sequence kernel is defined as

$$K_n(s,t) = \sum_{u \in \Sigma^\star} \sum_{\mathbf{i}:s[\mathbf{i}]=u} \sum_{\mathbf{j}:t[\mathbf{j}]=u} \lambda^{(i[n]-i[1]+1)+(j[n]-j[1]+1)}$$

**Equation 1.1:** Sequence Kernel

where $\lambda$ is a kernel parameter that can be thought of as a gap penalty, **i** refers to a vector of length *n* that consists of the indices of string *s* that correspond to the subsequence *u*. And, the value $i[n] - i[1] + 1$ can be regarded as the total length of the span of *s* that constitutes a particular occurrence of the subsequence *u*. Following Rousu *et al.* (2005), they combine the kernels of orders one through four through an exponential weighting,

$$K(s,t) = \sum_{i=1}^{N} \mu^{1-i} K_i(s,t)$$

**Equation 1.2:** Combining Sequential Kernels of Different Order

**Parts of Speech**

Parts of Speech information is most commonly exploited in all NLP tasks. One of the most important reasons is that they provide a crude form of word sense disambiguation.

**Adjectives only**

Adjectives have been used most frequently as features amongst all parts of speech. A strong correlation between adjectives and subjectivity has been found. Although all the parts of speech are important people most commonly used adjectives to depict most of the sentiments and a high accuracy have been reported by all the works concentrating on only adjectives for feature generation. Pang Lee *et al.* (2002) achieved as accuracy of around 82.8% in movie review domains using only adjectives in movie review domains.

|  | Proposed Word Lists |
|---|---|
| Human 1 | positive: dazzling, brilliant, phenomenal, excellent, fantastic |
|  | negative: suck, terrible, awful, unwatchable, hideous |

| Human 2 | positive: gripping, mesmerizing, riveting, spectacular, cool, awesome, thrilling, badass, excellent, moving, exciting |
|  | negative: bad, clichéd, sucks, boring, stupid, slow |

**Table 1.1:** Word List containing Positive and Negative Adjectives

**Adjective-Adverb Combination**

Most of the adverbs have no prior polarity. But when they occur with sentiment bearing adjectives, they can play a major role in determining the sentiment of a sentence. Benamara *et al.* (2007) have shown how the adverbs alter the sentiment value of the adjective that they are used with. Adverbs of degree, on the basis of the extent to which they modify this sentiment value, are classified as:

- Adverbs of affirmation: certainly, totally
- Adverbs of doubt: maybe, probably
- Strongly intensifying adverbs: exceedingly, immensely
- Weakly intensifying adverbs: barely, slightly
- Negation and minimizers: never

The work defined two types of AACs:
1. Unary AACs: Containing one adverb and one adjective. The sentiment score of the adjective is modified using the adverb adjoining it.
2. Binary AACs: Containing more than one adverb and an adjective. The sentiment score of the AAC is calculated by iteratively modifying the score of the adjective as each adverb gets added to it. This is equivalent to defining a binary AAC in terms of two unary AACs iteratively defined.

To calculate the sentiment value of an AAC, a score is associated with it based on the score of the adjective and the adverb. Certain axiomatic rules are specified to specify the way the adverbs modify the sentiment of the adjective. One such axiom can be stated as:

'Each weakly intensifying adverb and each adverb of doubt has a score less than or equal to each strongly intensifying adverb / adverb of affirmation.'

Then, certain functions are described in order to quantify the axioms. A function f takes an adjective-adverb pair and returns its resultant score.

1. *Affirmative and strongly intensifying adverbs*

- AAC-1. If $sc(adj) > 0$ and $adv \in AFF \cup STRONG$, then $f(adv, adj) \geq sc(adj)$.
- AAC-2. If $sc(adj) < 0$ and $adv \in AFF \cup STRONG$, then $f(adv, adj) \leq sc(adj)$.

For example, f value for 'immensely good' is more positive than the score for the positive adjective 'good'.

  2. *Weakly intensifying adverbs*

- AAC-3. If $sc(adj) > 0$ and $adv \in WEAK$, then $f(adv, adj) \leq sc(adj)$.
- AAC-4. If $sc(adj) < 0$ and $adv \in WEAK$, then $f(adv, adj) \geq sc(adj)$.

For example, f value for 'barely good' is more negative than the score for the positive adjective 'good'. This is the effect that the weakly intensifying adverb has.

  3. *Adverbs of doubt*

- AAC-5. If $sc(adj) > 0$, $adv \in DOUBT$, and $adv' \in AFF \cup STRONG$, then $f(adv, adj) \leq f(adv', adj)$.
- AAC-6. If $sc(adj) < 0$ is negative, $adv \in DOUBT$, and $adv' \in AFF \cup STRONG$, then $f(adv, adj) \geq f(adv', adj)$.

For example, f value for 'probably good' is less than 'immensely good'

  4. *Minimizers*

AAC-7. If $sc(adj) > 0$ and $adv \in MIN$, then $f(adv, adj) \leq sc(adj)$.
- AAC-8. If $sc(adj) < 0$ and $adv \in MIN$, then $f(adv, adj) \geq sc(adj)$.

For example, 'hardly good' is less positive than the positive adjective 'good'

**Scoring algorithms**

a. *Variable scoring algorithm:*

The algorithm modifies the score of the AAC using the function f defined as follows:

- If $adv \in AFF \cup STRONG$, then:

  $$f_{VS}(adv, adj) = sc(adj) + (1 - sc(adj)) \times sc(adv)$$

  if $sc(adj) > 0$. If $sc(adj) < 0$,

  $$f_{VS}(adv, adj) = sc(adj) - (1 - sc(adj)) \times sc(adv).$$

- If $adv \in WEAK \cup DOUBT$, VS reverses the above and returns

  $$f_{VS}(adv, adj) = sc(adj) - (1 - sc(adj)) \times sc(adv)$$

  if $sc(adj) > 0$. If $sc(adj) < 0$, it returns

  $$f_{VS}(adv, adj) = sc(adj) + (1 - sc(adj)) \times sc(adv).$$

**Algorithm 1.1**: Scoring Algorithm for Adjective Adverb Combination

Thus, the resultant score of the AAC is the score of the adjective which is suitably adjusted with the effect of the adverb.

b. *Adjective priority scoring algorithm:*

- If $adv \in AFF \cup STRONG$, then

  $$f_{APS^r}(adv, adj) = \min(1, sc(adj) + r \times sc(adv)).$$

  if $sc(adj) > 0$. If $sc(adj) > 0$,

  $$f_{APS^r}(adv, adj) = \min(1, sc(adj) - r \times sc(adv)).$$

- If $adv \in WEAK \cup DOUBT$, then $APS^r$ reverses the above and sets $f_{APS^r}(adv, adj) = \max(0, sc(adj) - r \times sc(adv))$. if $sc(adj) > 0$. If $sc(adj) < 0$, then $f_{APS^r}(adv, adj) = \max(0, sc(adj) + r \times sc(adv))$.

They give priority to adjectives over the adverbs and modify the score of the adjective by a weight r. This weight r decides the extent to which an adverb influences the score of an adjective.

c. *Adverb priority scoring algorithm*

- If $adv \in AFF \cup STRONG$, then

  $$f_{AdvFS^r}(adv, adj) = \min(1, sc(adv) + r \times sc(adj))$$

  if $sc(adj) > 0$. If $sc(adj) < 0$,

  $$f_{AdvFS^r}(adv, adj) = \max(0, sc(adv) - r \times sc(adj)).$$

**Topic-Oriented Features**

Bag-of-words and phrases are extensively used as features. But in many domains, individual phrase values bear little relation with overall text sentiment. A challenge in Sentiment

Analysis of a text is to exploit those aspects of the text which are in some way representative of the tone of the whole text. Often misleading phrases (thwarted expectations) are used to reinforce sentiments. Using bag-of-words or individual phrases will not be able to distinguish between what is said locally in phrases and what is meant globally in the text like drawing of contrasts between the reviewed entity and other entities, sarcasm, understatement, and digressions, all of which are used in abundance in many discourse domains. These features were developed by Turney *et al.* (2002).

1  Semantic Orientation with PMI

*Semantic Orientation* (SO) refers to a real number measure of the positive or negative sentiment expressed by a word or phrase. The *value phrases* are phrases that are the source of SO values. Once the desired value phrases have been extracted from the text, each one is assigned an SO value. The SO of a phrase is determined based upon the phrase's *pointwise mutual information* (PMI) with the words "excellent" and "poor". PMI is defined by Church and Hanks (1989) as follows:

$PMI(w_1, w_2) = \log_2(p(w_1 \ \& \ w_2)/p(w_1)p(w_2))$.

The SO for a phrase is the difference between its PMI with the word "excellent" and its PMI with the word "poor." i.e

SO (phrase) = PMI (phrase, "*excellent*") - PMI (phrase, "*poor*")

Intuitively, this yields values above zero for phrases with greater PMI with the word "excellent" and below zero for greater PMI with "poor". A SO value of zero would indicate a completely neutral semantic orientation.

|   | **First Word** | **Second Word** | **Third Word (Not Extracted)** |
|---|---|---|---|
| 1. | JJ | NN or NNS | anything |
| 2. | RB, RBR or RBS | JJ | not NN nor NNS |
| 3. | JJ | JJ | not NN nor NNS |
| 4. | NN or NNS | JJ | not NN or NNS |
| 5. | RB, RBR or RBS | VB, VBD, VBN or VBC | anything |

**Table 1.2:** Phrase Patterns Used for Extracting Value Phrases - Turney (2002)

**Osgood semantic differentiation with WordNet**

WordNet relationships are used to derive three values pertinent to the emotive meaning of adjectives. The three values correspond to the *potency* (strong or weak), *activity* (active or passive) and the *evaluative* (good or bad) factors introduced in Charles Osgood's Theory of Semantic Differentiation (Osgood *et al.*, 1957). These values are derived by measuring the relative minimal path length (MPL) in WordNet between the adjective in question and the pair of words appropriate for the given factor. In the case of the *evaluative* factor (EVA) for example, the comparison is between the MPL between the adjective and"good" and the MPL between the adjective and"bad". The values can be averaged over all the adjectives in a text, yielding three real valued feature values for the text.

"Sentiment expressed with regard to a particular subject can best be identified with reference to the subject itself", Natsukawa and Yi (2003).

In some application domains, it is known in advance what the topic is toward which sentiment is to be evaluated. This can be *exploited* by creating several classes of features based upon the SO values of phrases given their position in relation to the topic of the text. In opinionated texts there is generally a single primary subject about which the opinion is favorable or unfavorable. But secondary subjects are also useful to some extent.

*Ex: Opinion (reference) to author in a book review may be useful in a book review.*
*Ex: In a product review, the attitude towards the company which manufactures the product may be pertinent.*

The work considers the following classes of features:

    a. *Turney Value*

The average value of all value phrases' SO values for the text

    b. *In sentence with THIS WORK*

The average value of all value phrases which occur in the same sentence as a reference to the work being reviewed

    c. *Following THIS WORK*

The average value of all value phrases which follow a reference to the work being reviewed directly, or separated only by the copula or a preposition

    d. *Preceding THIS WORK*

The average value of all value phrases which precede a reference to the work being reviewed directly, or separated only by the copula or a preposition

  e. *In sentence with THIS ARTIST*

With reference to the artist

  f. *Following THIS ARTIST*

With reference to the artist

  g. *Preceding THIS ARTIST*

With reference to the artist

  h. *Text-wide EVA*

The average EVA value of all adjectives in a text

  i. *Text-wide POT*

The average POT value of all adjectives in a text

  j. *Text-wide ACT*

The average ACT value of all adjectives in a text

  k. *TOPIC-sentence EVA*

The average EVA value of all adjectives which share a sentence with the topic of the text

  l. *TOPIC-sentence POT*

The average POT value of all adjectives which share a sentence with the topic of the text

  m. *TOPIC-sentence ACT*

The average ACT value of all adjectives which share a sentence with the topic of the text

## 1.5   MACHINE LEARNING APPROACHES

In his work, Pang Lee *et al.* (2002, 2004), compared the performance of Naïve Bayes, Maximum Entropy and Support Vector Machines in SA on different features like considering only unigrams, bigrams, combination of both, incorporating parts of speech and position information, taking only adjectives *etc*. The result has been summarized in the Table 1.3.

It is observed from the results that:
 a. Feature presence is more important than feature frequency.
 b. Using Bigrams the accuracy actually falls.
 c. Accuracy improves if all the frequently occurring words from all parts of speech are taken, not only Adjectives.

d. Incorporating position information increases accuracy.
   e. When the feature space is small, Naïve Bayes performs better than SVM. But SVM's perform better when feature space is increased.

When feature space is increased, Maximum Entropy may perform better than Naïve Bayes but it may also suffer from overfitting.

| **Features** | **Number of Features** | **Frequency or Presence?** | **NB** | **ME** | **SVM** |
|---|---|---|---|---|---|
| Unigrams | 16165 | Freq. | **78.7** | N/A | 72.8 |
| Unigrams | 16165 | Pres. | 81.0 | 80.4 | **82.9** |
| Unigrams+bigrams | 32330 | Pres. | 80.6 | 80.8 | **82.7** |
| Bigrams | 16165 | Pres. | 77.3 | **77.4** | 77.1 |
| Unigrams+POS | 16695 | Pres. | 81.5 | 80.4 | **81.9** |
| Adjectives | 2633 | Pres. | 77.0 | **77.7** | 75.1 |
| Top 2633 unigrams | 2633 | Pres. | 80.3 | 81.0 | **81.4** |
| Unigrams+position | 22430 | Pres. | 81.0 | 80.1 | **81.6** |

**Table 1.3:** Accuracy Comparison of Different Classifiers in SA on Movie Review Dataset

Bikel *et al.* (2007) implemented a Subsequence Kernel based Voted Perceptron and compares its performance with standard Support Vector Machines. It is observed that as the number of true positives increase, the increase in the number of false positives is much less in Subsequence Kernel based voted Perceptrons compared to the bag-of-words based SVM's where the increase in false positives with true positives is almost linear. Their model, despite being trained only on the extreme one and five star reviews, formed an excellent continuum over reviews with intermediate star ratings, as shown in the figure below. The authors comment that "It is rare that we see such behavior associated with lexical features which are typically regarded as discrete and combinatorial. Finally, we note that the voted perceptron is making distinctions that humans found difficult…".

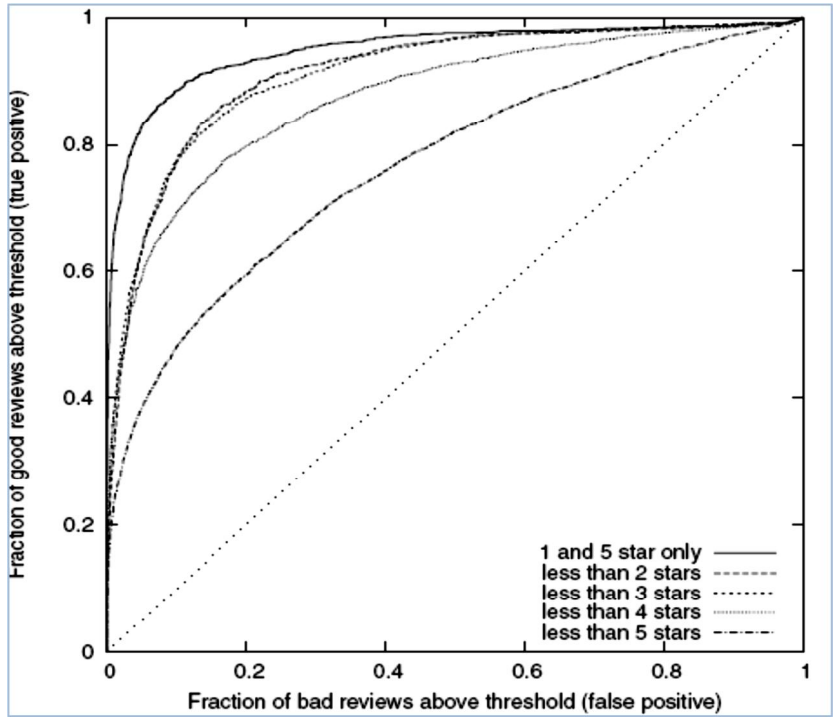

**Figure 1.1**: Ratio of True Positives and False Positives using Subsequence Kernel based Voted Perceptrons

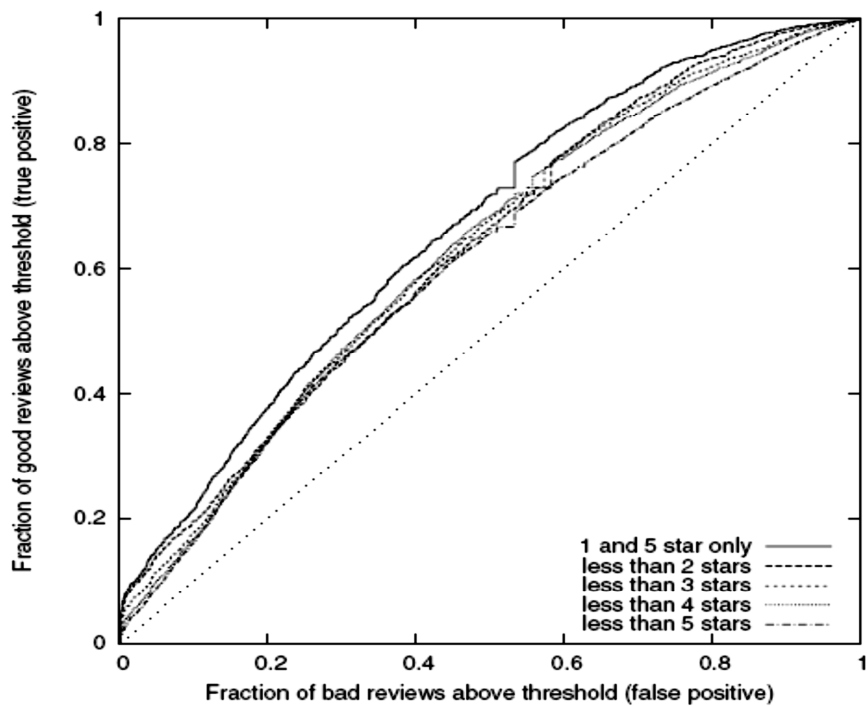

**Figure 1.2:** Ratio of True Positives and False Positives using Bag-of-Features SVM

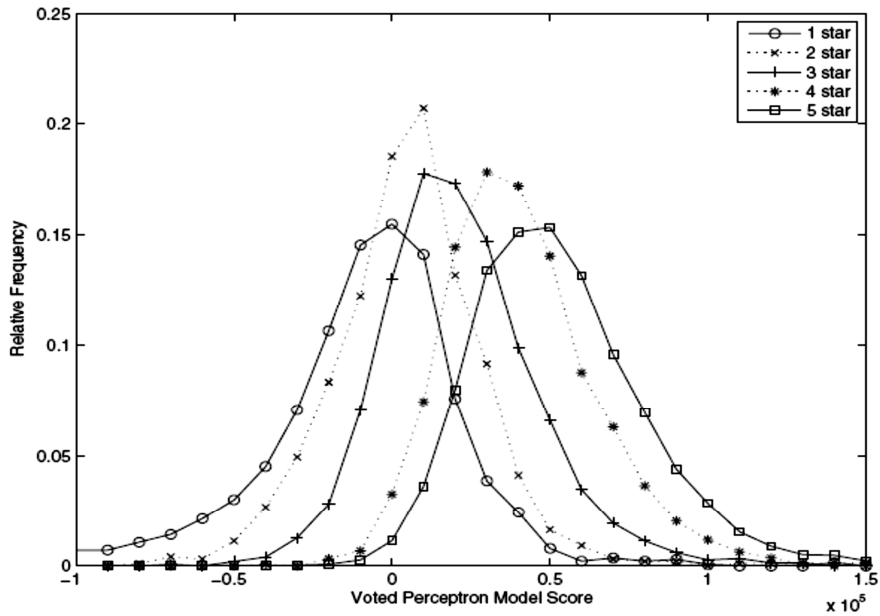

**Figure 1.3:** Distribution of Voted Perceptron Model Scores by Number of Stars

Pande, Iyer *et al.* performs a detailed comparison of the different classifiers in two phases under two settings. In phase 1, the classifiers are made to distinguish between subjective and objective documents. In phase 2, the classifiers are made to classify positive from negative documents filtered by phase 1. In each phase classifiers have been tested without and with boosting to enhance performance. The classifiers tested are Bayesian Logistic Regression (BLR) with Gaussian and Laplacian prior, Naïve Bayes, Support Vector Machines with linear, polynomial and radial basis functions kernels and Voted Perceptrons.

| Classifier | Avg F1 | Avg Acc. | Max Acc. |
|---|---|---|---|
| Naïve Bayes | 0.4985 | 78.66 | 81.55 |
| SVM(Linear) | 0.4700 | 80.83 | 83.42 |
| SVM(Poly) | 0.4536 | 79.71 | 81.97 |
| SVM(RBF) | 0.0864 | 78.05 | 79.4 |
| Voted Perceptron | 0.4300 | 78.64 | 83.66 |

**Table 1.4:** Accuracy Comparison of Different Classifiers, without Boosting (Skewed Dataset)

| Classifier | Avg F1 | Avg Acc. | Max Acc. |
|---|---|---|---|
| BLR(Gauss) | 0.515 | 80.09 | 85.51 |
| BLR(Laplace) | 0.510 | 79.98 | 87.04 |
| Naïve Bayes | 0.563 | 75.63 | 84.18 |
| SVM(Linear) | 0.557 | 80.2 | 85.44 |
| SVM(Poly) | 0.499 | 79.22 | 83.37 |
| Voted Perceptron | 0.508 | 78.70 | 83.62 |

**Table 1.5:** Accuracy Comparison of Different Classifiers, with Boosting (Skewed Dataset)

| Classifier | Avg F1 | Avg Acc. | Max Acc. |
|---|---|---|---|
| Naïve Bayes | 0.882 | 83.91 | 84.80 |
| SVM(Linear) | 0.880 | 83.29 | 85.736 |
| SVM(Poly) | 0.8571 | 80.80 | 82.946 |
| SVM(RBF) | 0.757 | 68.22 | 74.573 |
| Voted Perceptron | 0.875 | 82.96 | 85.43 |

**Table 1.6:** Accuracy Comparison of Different Classifiers, without Boosting (Balanced Dataset)

In phase 1, it is observed that:

1. Accuracy very high but F1 scores are low due to skewed data (number of objective samples were 10,000 whereas number of polar samples were 25000) due to which either recall is low or false-positive rate is considerably high.
2. Without boosting, undersampling or oversampling maximum recall attained was 65%.
3. Lowest recall was attained for feature based on Gain Ratio and/or SVM with RBF kernel.
4. With boosting highest recall attained was 85% with false negative rate 30%.
5. Naïve Bayes and SMO gave the best recall.
6. Voted Perceptrons, BLR gave less than 65% recall
7. Voted Perceptrons and SVM(linear) give the best accuracy.
8. It is astonishing to see SVM(RBF) give the lowest accuracy. The reason may be overfitting.

In phase 2, it is observed that:

1. SVM with RBF again performs badly.
2. SVM (linear) and Voted Perceptrons again have the best accuracy. They are found to give good results with both Information gain, Chi-Square feature selection methods.
3. Naïve Bayes favors Chi-Square over Information Gain.
4. Boosting does not improve performance much.

Mullen *et al.* (2004) used Support Vector Machines with diverse information measures by using features like the PMI, Lemma, Turney, Osgood values along with other topic oriented features.

It is observed that:

1. Using only Turney values, a high accuracy can be achieved.
2. The addition of Osgood values does not seem to yield improvement in any of the models.
3. Using only Lemmas instead of Unigrams result in a much better performance.

4. The inclusion of all PMI values with lemmas outperforms the use of only the Turney values, suggesting that the incorporation of the available topic relations is helpful.
5. The best performance is achieved by using Lemmas and PMI or Osgood Values.

| Model | 5 folds | 10 folds | 20 folds | 100 folds |
|---|---|---|---|---|
| Turney Values only | 72% | 73% | 72% | 72% |
| All (THIS WORK and THIS ARTIST) PMI | 70% | 70% | 68% | 69% |
| THIS WORK PMI | 72% | 69% | 70% | 71% |
| All Osgood | 64% | 64% | 65% | 64% |
| All PMI and Osgood | 74% | 71% | 74% | 72% |
| Unigrams | 79% | 80% | 78% | 82% |
| Unigrams, PMI, Osgood | 81% | 80% | 82% | 82% |
| Lemmas | 83% | 85% | 84% | 84% |
| Lemmas and Osgood | 83% | 84% | 84% | 84% |
| Lemmas and Turney | 84% | 85% | 84% | 84% |
| Lemmas, Turney, text wide Osgood | 84% | 85% | 84% | 84% |
| Lemmas, PMI, Osgood | 84% | 85% | 84% | 86% |
| Lemmas and PMI | 84% | 85% | 85% | 86% |
| Hybrid SVM (PMI/Osgood and Lemmas) | 86% | 87% | 84% | 89% |

**Table 1.7:** Accuracy Comparison of Different Features on SVM using a Linear Kernel

## 1.6 COGNITIVE APPROACHES (DISCOURSE)

### 1.6.1 What is Subjectivity Analysis, Perspective and Narratives?

The input to the Sentiment Classifier is always an opinionated text *i.e.* a text containing positive or negative sentiment. Thus one needs to filter out objective facts from a text and subject only the opinions to the Sentiment Classifier. This work of extracting or filtering out the objective facts from subjective opinions is called Subjectivity Analysis.

A piece of text often contains other person's point of view or accounts from a third person which contain a gamut of emotions or opinions from different characters or perspectives. Thus it is important to identify the character to who an opinion can be attributed to. Thus the objective here is not only to detect whether a piece of text is opinionated or not but also who is responsible for that opinion.

A *narrative* is a story that is created in a constructive format. It describes a sequence of fictional and non-fictional events. A narrative can also be told by a character within a larger narrative. It is the fiction-writing mode whereby the narrator communicates directly to the reader.

A *perspective* is the point of view. Narrative is told from the perspective of one or more of its characters. It can also contain passages not told from the perspective of any character.

Subjective sentences *portray* a character's *thoughts* – represented thoughts or present a *scene perceived* by the character – represented perception, private state such as seeing, wanting or feeling ill – that is some perceptual, psychological or experiential state not open to objective observation or verification.

Objective sentences present the story *directly,* rather than through *thoughts or perceptions* of a character.

Thus subjective sentences take a character's psychological point of view or POV (Uspensky 1973). For narrative understanding it is absolutely essential to track the POV as it distinguishes between beliefs of characters and facts of the story.

### 1.6.2 Discourse-Level Analysis

But in order to identify the character and its perspective a sentence level analysis will not do. This demands a discourse level analysis as the sentences are not always explicitly marked by subjective elements and the subjective sentences do not directly imply the subjective character, Wiebe *et al.* (1991).

*Example: [1.1]He wanted to talk to Dennys. [1.2]How were they going to be able to get home from this strange desert land into which they had been cast and which was heaven knew where in all the countless solar systems in all the countless galaxies? [L'Engle, Many Waters, p. 91]*

Sentence (1.2) is a represented thought, and (2.2) is a represented perception, presenting what the character sees as he sees it, yet neither is explicitly marked as such. Also, neither indicates who the subjective character is.

Subjective Sentences which do no contain any subjective elements or subjective character appear in the midst of other subjective sentences attributed to the same subjective character. Thus subjectivity needs to be determined at discourse level. Instead of subjective or objective sentences we have subjective and objective contexts consisting of 1 or more subjective or objective sentences attributed to the same subjective character or objective sentences.

There are regularities in a way the author initiate, continue or resumes a character's POV. Certain combination of the sentence features (like tense, aspect, lexical items expressing subjectivity, identity of actors or experiencers of those states of affairs) and the current context (like whether the previous sentence was subjective or objective or whether there was a scene break or paragraph break) helps in detecting the continuity of the POV of a character.

The POV tracking needs an extensive discourse analysis. For example, the use of a full noun when a pronoun would have sufficed denotes that change in POV has occurred. On the other hand use of anaphoric pronoun denotes continuity in current POV.

The private state reports of a character, that expresses whether the subjective character is ill or angry, can only be reported. Now, if "*John was furious*" is the subjective statement of a character Mary, it is only her *represented thought or opinion* about John. Thus the distinction between private state report and represented thought is essential for discourse processing. This is because, as the subjective character is always the subject of a private state report, pronouns can be used to refer him despite references to some other entity of same number and gender. But in a represented thought the referent of a pronoun can be someone in an earlier represented thought.

Example:

"*Dwayne wasn't sure what John was scared of. What in the arcade could scare a boy like that? He could see tear's in John's eyes. He could tell they were tears because …. Maybe that was why he was crying.*

   *"I want to leave", he said."*

Here the last sentence is objective but the previous sentences are subjective sentences of Dwayne. 'He' in the last sentence refers to Dwayne even though John is the last previously mentioned entity. This suggests there is a change of POV and discourse analysis is required to detect it.

**1.6.3 Subjective Contexts**

Recognition of a subjective context requires the presence of linguistic signals. Wiebe *et al.* (1988) recognizes the following subjective signals..

a. Psychological Verbs, Actions, Adjectives and Perceptual Verbs

   1. Psychological verbs (e.g. 'think', 'wonder', 'realize', 'went')
   2. Perceptual verbs (e.g.. 'see', 'hear')
   3. Psychological adjectives (e.g., 'delighted', 'happy', 'jealous', 'scared')
   4. Psychological actions- (e.g.,"he smiled","she gasped","she winced"

   *a. Subjective Elements*

1. Exclamations, which express emotions
2. Questions, which express wonder

3. Epithets, such as 'the bastard', which express some qualification of the referent.
4. Kinship *terms,* e.g., 'Daddy', 'Morn', and 'Aunt Margaret', which express a relationship to the referee.
5. Evaluative adjectives, which express an attitude toward the referent, e.g., 'ghastly', 'surprising', 'poor', and 'damned',
6. Intensifiers such as 'too', 'quite', and 'so' are also evaluative
7. Emphasizers like"really","just"

### b. More Subjective Elements

1. Modal verbs of obligation, possibility, and necessity. For example, 'should' is a modal verb of obligation)
2. *content (or attitudinal) disjuncts* which comment on the content of the utterance. For example,"*likely',* 'maybe', 'probably', and 'perhaps' express some degree of doubt
3. Conjuncts, which comment on the connection between items. For example, 'anyhow', 'anyway', 'still', and 'after all' express concession
4. Uses of 'this', 'that', 'these', and 'those' that Robin Lakoff (1974) has identified as *emotional deixis.*

*In* conversation, they are"generally linked to the speaker's emotional involvement in the subject-matter of his utterance" (Lakoff 1974: 347); in thlrd-person narrative, they are linked to the subjective character's emotional involvement in the subject matter of his thoughts or perceptions.

Examples:

*[2.1]She [Hannah]* **winced** *as she heard them crash to the platform.*

*[3.1]He could tell they were tears because his eyes were* **too** *shiny. Too round.*

*[4.1]Jody managed a frail smile. [4.2]She was* **a little** *bit ashamed. [4.3] She should* **really** *try to be more cheerful for Aunt Margaret's sake. [4.4]***After all**, *Aunt Margaret had troubles of her own--she was the mother of that ghastly Dill.*

In the above examples, the words in bold are subjective elements. The presence of a subjective element indicates the presence of a subjective character.

### 1.6.4   Identifying a Subjective Character

Subjective Character can sometimes be directly identified in a sentence (for example when there is a narrative parenthetical), Wiebe *et al.* (1990).

If it is not identifiable, then it is one of the 2 previously mentioned characters-

1. the subjective character of the previous subjective sentence

It either continues the POV of the subjective character or resumes it.

2. actor of an action denoted by a previous objective sentence

Examples:

*[5.1]Why, Jake,you lazy bean," Augustus said, [5.2] and walked off. [5.3] Jake had a stubborn streak in him, [5.4]and once it was activated even Call could seldom do much with him.*

(5.3) and (5.4) represent the point of view of Augustus, the actor of an action denoted by a previous objective sentence, (5.1). But the last subjective character is Jake, so Augustus's point of view is initiated, not merely resumed or continued.

In order to identify the subjective character one needs to keep track of expected subjective characters encountered. However drastic spatial and temporal discontinuities can block the continuation or resumption of a character's POV.

Ex: *scene break, paragraph break*

The subjective character may also be identified from a private-state sentence. It *can* be the experiencer of a private state sentence. **Exception** occurs when a subjective sentence is followed by a private state report without paragraph break where the experiencer is different from the subjective character.

Example of a scene break:

*[6.l]Drown me?" Augustus said. [6.2] Why if' anybody had tried it, those girls would have clawed them to shreds." [6.3] He knew Call was mad, [6.4] but wasn't much inclined to humor him. [6.5] It was his dinner table as much as Call's, [6.6] and if Call didn't like the conversation he could go to bed.*

*[6.7] Call knew there was no point in arguing. [6.8] That was what Augustus wanted: argument. [6.9] He didn't really care what the question was, [6.10] and it made no great difference to him which side he was on. [6.11] He just plain loved to argue.*

Sentences (6.1)-(6.2) are Augustus's subjective sentences and (6.7)-(6.11) are Call's. So, (6.7) initiates a new point of view. It is a private-state sentence and the subjective character, Call, is the experiencer of the private state denoted.

Subjective character of a private state report is always the experiencer which can be directly determined from the sentence. But such cannot be said in case of a represented thought. If private-state report is indicated to be a represented thought, then subjective character is the expected subjective character. But an important aspect is the scope of the subjective element. Consider the following examples.

*[7.1] Japheth, **evidently** realizing that they were no longer behind him, turned around [7.2]and jogged back toward them, seemingly cool and unwinded.*

*[8.1]Urgh! She) the [girl] thought. [8.2]How could the poor thing have married him in the first place?*

*[8.3]Johnnie Martin could not believe that he was seeing that old bag's black eyes sparkling with disgust and unsheathed contempt at him.*

Sentence (8.3) is a private-state report and the subjective character is the experiencer (Johnnie Martin). This is so even though (8.3) contains the subjective element 'old bag' and even though there is an expected subjective character (the girl) when it is encountered. Because 'old bag' appears within the scope of the private-state term 'believe', it is not considered in identifying the subjective character. On the other hand, the subjective clement 'evidently ' in (7.1) is not in the scope of 'realizing' (*i.e.*, it is *non-subordinated),* so it can be used to identify the subjective character.

---

*if the sentence contains a narrative parenthetical then*
    *SC is the subject of the parenthetical*
*else if the sentence is a private-state sentence then*
    *if it has a non-subordinated subjective clement*
    *or the text situation is continuing-subjective then*
        *SC is identified from the previous context*
    *else SC is the experiencer*
    *end if*
*else*
    *SC is identified from the previous context*
*end if*

---

**Algorithm 1.2**: To Identify the Subjective Character

***if*** *there are two expected subjective characters then*

    ***if*** *the sentence is about the last active character then*

        *SC is the last subjective character*

    ***else*** *SC is the last active character*

    ***end if***

***else if*** *there is an expected subjective character then*

    *SC is the expected subjective character*

***else*** *SC is unidentified*

***end if***

**Algorithm 1.3**: To Identify the Subjective Character from Previous Context

### 1.6.5 Identifying Perspective in Narrative

A belief space is accessed by a stack of individuals. It consists of what the bottom member of the stack believes that what the top member believes.

The reader is always the bottom member. The belief space corresponding to a stack consisting only of the reader contains the set of propositions that the reader believes are true. The CP determines the current belief space with respect to which references are understood.

***if*** *'X' is an indefinite noun phrase of the form 'a Y' then*

    *create a new concept, N; build in CP's belief space the proposition that N is a Y;*

    *return N*

***else if*** *'X' is a definite noun phrase or proper name,*

    *if a proposition that N is X can be found in the CP's belief space,*

        *return N*

    *else if a proposition that N is X can be found in a belief space other than the CP's, then add the found proposition to the CP's belief space;*

        *return N*

    *else create a new concept, N; build in CP's belief space the proposition that N is X;*

*return N*

**Algorithm 1.4**: Algorithm for Understanding a Non-Anaphoric, Specific Reference 'X' in Third-Person Narrative

If a reference is a subjective element, such as 'the bastard', it cannot be understood entirely propositionally, since it expresses subjectivity. How it should be understood depends on the particular subjective element.

It does not understand anaphoric references. However, anaphor comprehension can be affected by perspective.

Examples:

*[9.1]The man had turned. [9.2]He started to walk away **quickly** in the direction of the public library.*
*[9.3]"O.K.," said Joe,"get Rosie."*
*[9.4]Zoe crept back to the blinker. [9.5]She felt hollow in her stomach. [9.6]She'd never **really** expected to see the Enemy again. [Ones], War Work, p. 64]*

'The Enemy' is an anaphoric reference that occurs in a subjective context (established by (9.5), which is a Psychological report). It co-specifies 'the man' in (9.1) and 'He' in (9.2). It reflects Zoe's belief that the man is an enemy spy, although it is not at all clear to the reader, at point' that he is.

Personal pronouns can also reflect the beliefs of a character.

Assertive indefinite pronouns ex. 'someone', 'something', 'somebody' refer to particular people, things, *etc*., without identifying them. When referring to a particular referent, a speaker typically uses an assertive indefinite pronoun if

(1) she doesn't know the identity of the referent
(2) She doesn't want the addressee to know the identity of the referent, or
(3) she doesn't believe that the identity of the referent is relevant to the conversation.

A character's thoughts end perceptions are not directed toward an addressee, and so the first of these uses is the predominant one in subjective contexts.

*Example:[10] Suddenly she [Zoe] gasped. She had touched **somebody**! [O'Neal, War Work, p. 129]*

There is no explicit statement in the novel that Zoe does not know whom she touched; this has to be inferred from the use of 'somebody'.

Definite references are used only if the speaker believes that the addressee has enough information to interpret them. Specific indefinite references are used in a subjective context when the referent is unfamiliar to the subjective character. However, the referent may not be unknown to the reader or to the other characters.

*[11] There they [the King and his men] saw close beside them a great rubbleheap; and suddenly they were aware of **two small figures** lying on it at their ease, grey-clad, hardly to be seen among the stones. [Tolkien, The Two Towers, p. 206]*

The reader knows that the King and his men have come upon two hobbits, Merry end Pippin. The King and his men do not know the hobbits, but other characters also present in the scene do know them. When the King and his man are on the top of the CP (after 'saw' and continued by 'were aware of'), the hobbits are not referred to by name, but as 'two small figures'. Thus new referents are created and propositions are built in the belief space of the King and his men that they are small figures. The new referents can be asserted to be co-extensional with the concepts who the reader and other characters believe are named 'Merry' and 'Pippin'.

### 1.6.6 Evaluation

The algorithms were tested on 450 sentential input items (exclusive of paragraph and scene breaks) from each of two novels, *Lonesome Dove* by Larry McMurtry and *The Magic of the Glits* by Carole S. Adler. *Lonesome Dove* is an adult novel that has many subjective characters, and *The Magic of the Glits* is a children's' novel that has one main subjective character. The input items are those of the complete sentences of every fifth page of these novels, starting in *Lonesome Dove* with page 176 and ending with page 236 (13 pages total), and starting in *The Magic of the Glits* with page 1 and ending with page 86 (18 pages total). (For each book, the first part of an additional page was used to make the number of input items exactly equal to 450.) Page 176 in *Lonesome Dove* is the beginning of a chapter in the middle of the novel. The earlier pages of the novel were considered during the development of the algorithm.

| Interpretation | Actual Instances | Primary Errors | Incorrect Interpretations |
|---|---|---|---|
| <subjective, x> | 271/450 (60%) | 20/271 (7%) | 13 objective<br>7 (subjective, y), y ≠ x |
| Objective | 179/450 (40%) | 7/179 (4%) | 7 (subjective, x) |
| Objective, other than simple quoted speech | 54/450 (12%) | 7/54 (13%) | 7 (subjective, x) |

**Table 1.8:** Results for Lonesome Drove by Interpretation

| Point-of-View Operation | Actual Instances | Primary Errors | Incorrect Interpretations |
|---|---|---|---|
| Continuation | 215/450 (48%) | 11/215 (5%) | 1 initiation<br>10 objective |
| Resumption | 20/450 (4%) | 0/20 (0%) | - |
| Initiation | 36/450 (8%) | 9/36 (25%) | 5 resumptions<br>1 initiation<br>3 objective |
| Objective | 179/450 (40%) | 7/179 (4%) | 4 continuations<br>3 resumptions |
| Objective, other than simple quoted speech | 54/450 (12%) | 7/54 (13%) | 4 continuations<br>3 resumptions |

**Table 1.9:** Results for Lonesome Drove by Point-of-View Operation

*In Lonesome Dove,* out of the 450 input items, the algorithm committed 27 primary errors (6%) and 28 secondary errors (6%). Many of the input items, 125 of them (28%), are simple items of quoted speech (*i.e.*, they do not have potential subjective elements in the discourse parenthetical, or subordinated clauses outside the quoted string that have private-state terms, private-state-action terms, or potential subjective elements).

      In Table 2.8, the first row, is interpreted as: Out of the 271 actual subjective sentences, the algorithm committed 20 primary errors. It interpreted 13 subjective sentences to be objective, and 7 to be the subjective sentence of the wrong subjective character.

In Table 2.9, the first row, is interpreted as: Out of the 215 items that actually continue a character's point of view, the algorithm committed 11 primary errors. It interpreted 1 of them to be an initiation and 10 to be objective. Notice that the last column of the row for initiations includes an initiation. This means that for one actual initiation, the algorithm was correct that a character's point of view was initiated, but incorrect as to the identity of that character.

| Interpretation | Actual Instances | Primary Errors | Incorrect Interpretations |
|---|---|---|---|
| <subjective, x> | 125/450 (28%) | 12/125 (10%) | 10 objective<br>2 (subjective, y), y ≠ x |
| Objective | 325/450 (72%) | 22/325 (7%) | 22 (subjective, x) |
| objective, other than simple quoted speech | 97/450 (22%) | 22/97 (23%) | 22 (subjective, x) |

**Table 1.10**: Results for The Magic of the Glits by Interpretation

| Point-of-View Operation | Actual Instances | Primary Errors | Incorrect Interpretations |
|---|---|---|---|
| Continuation | 79/450 (18%) | 4/79 (5%) | 4 objective |
| Resumption | 41/450 (9%) | 7/41 (17%) | 2 initiations<br>5 objective |
| Initiation | 5/450 (1%) | 1/5 (20%) | 1 objective |
| Objective | 325/450 (72%) | 22/325 (7%) | 4 continuations<br>9 resumptions<br>9 initiations |
| objective, other than simple quoted speech | 97/450 (22%) | 22/97 (23%) | 4 continuations<br>9 resumptions<br>9 initiations |

**Table 1.11**: Results for The Magic of the Glits by Point-of-View Operations

In *The Magic of the Glits,* out of the 450 input items, the algorithm committed 34 primary errors (8%) and 21 secondary errors (5%). There are 228 items that are simple quoted speech (51%). Tables 2.10 and 11 present the kinds of results for this novel that were given above in Tables 2.8 and 2.9 for *Lonesome Dove.*

## 1.7 SENTIMENT ANALYSIS AT IIT BOMBAY

Balamurali *et al.* (2011) presents an innovative idea to introduce sense based sentiment analysis. This implies shifting from lexeme feature space to semantic space *i.e.* from simple words to their synsets. The works in SA, for so long, concentrated on lexeme feature space or identifying relations between words using parsing. The need for integrating sense to SA was the need of the hour due to the following scenarios, as identified by the authors:

a. A word may have some sentiment-bearing and some non-sentiment-bearing senses
b. There may be different senses of a word that bear sentiment of opposite polarity
c. The same sense can be manifested by different words (appearing in the same synset)

Using sense as features helps to exploit the idea of sense/concepts and the hierarchical structure of the WordNet. The following feature representations were used by the authors and their performance were compared to that of lexeme based features:

a. A group of word senses that have been manually annotated (M)
b. A group of word senses that have been annotated by an automatic WSD (I)

c.  A group of manually annotated word senses and words (both separately as features) (Sense + Words(M))

d.  A group of automatically annotated word senses and words (both separately as features) (Sense + Words(I))

Sense + Words(M) and Sense + Words(I) were used to overcome non-coverage of WordNet for some noun synsets.

The authors used synset-replacement strategies to deal with non-coverage, in case a synset in test document is not found in the training documents. In that case the target unknown synset is replaced with its closest counterpart among the WordNet synsets by using some metric. The metrics used by the authors were LIN, LESK and LCH.

SVM's were used for classification of the feature vectors and IWSD was used for automatic WSD. Extensive experiments were done to compare the performance of the 4 feature representations with lexeme representation. Best performance, in terms of accuracy, was obtained by using sense based SA with manual annotation (with an accuracy of 90.2% and an increase of 5.3% over the baseline accuracy) followed by Sense(M), Sense + Words(I), Sense(I) and lexeme feature representation. LESK was found to perform the best among the 3 metrics used in replacement strategies.

One of the reasons for improvements was attributed to feature abstraction and dimensionality reduction leading to noise reduction. The work achieved its target of bringing a new dimension to SA by introducing sense based SA.

Aditya *et al.* (2010) introduced Sentiment Analysis to Indian languages namely, Hindi. Though, much work has been done in SA in English, little or no work has been done so-far in Hindi. The authors exploited 3 different approaches to tackling SA in Hindi:

1. They developed a sense annotated corpora for Hindi, so that any supervised classifier can be trained on that corpora.

2. They translated the Hindi document to English and used a classifier trained in English documents to classify it.

3. They developed a sentiment lexicon for Hindi called the Hindi SentiWordNet (H-SWN).

The authors first tried to use the in-language classifier and if training data was not available, they settled for a rough Machine Translation of the document to resource-rich English. In case MT could not be done they used the H-SWN and used a majority voting approach to find the polarity of the document.

As expected, the classifier trained in in-language documents gave the best performance. This was followed by the MT system of converting the source document to English. One of the reasons for its degraded performance can be attributed to the absence of Word Sense Disambiguation, which led to a different meaning of the Hindi word in English after translation. The lexical resource based approach gave the worse performance amongst the 3 approaches. This was mainly due to the coverage of the H-SWN which was quite low.

Aditya *et al.* (2010) took Sentiment Analysis to a new terrain by introducing it to micro-blogs namely, Twitter. They developed the C-Feel-It system that extracted posts called Tweets from the Twitter, related to the user query, and evaluated its sentiment based on 4 lexical resources. Twitter is a very noisy medium where the user posts different forms of slangs, abbreviations, smileys *etc*. There is also a high occurrence of spams generated by bots. Due to these reasons, the accuracy of the system deteriorated mainly because the words in the post were not present in the lexical resources. However, the authors used some form of normalization to compensate somewhat for the inherent noise. They also used a slang and emoticon dictionary for polarity evaluation. The authors used the 4 lexical resources *Taboada, Inquirer, SentiWordNet and Subjectivity Lexicon* and settled for a majority voting for the final polarity of the tweet. This work is novel mainly because it exploits a new domain.

## 1.8 DISCOURSE SPECIFIC SENTIMENT ANALYSIS

Marcu (2000) discussed probabilistic models for identifying elementary discourse units at clausal level and generating trees at the sentence level, using lexical and syntactic information from discourse-annotated corpus of RST. Wellner *et al.* (2006) considered the problem of automatically identifying arguments of discourse connectives in the PDTB. They modeled the problem as a predicate-argument identification where the predicates were discourse connectives and arguments served as anchors for discourse segments. Wolf *et al.* (2005) present a set of discourse structure relations and ways to code or represent them. The relations were based on Hobbs (1985). They report a method for annotating discourse coherent structures and found different kinds of crossed dependencies.

In the work, *Contextual Valence Shifters* (Polanyi *et al.,* 2004), the authors investigate the effect of *intensifiers, negatives, modals* and *connectors* that changes the prior polarity or valence of the words and brings out a new meaning or perspective. They also talk about pre-suppositional items and irony and present a simple weighting scheme to deal with them.

Somasundaran *et al*. (2009) and Asher *et al*. (2008) discuss some discourse-based supervised and unsupervised approaches to opinion analysis. Zhou *et al.* (2011) present an approach to identify discourse relations as identified by RST. Instead of depending on cue-phrase based methods to identify discourse relations, they leverage it to adopt an unsupervised approach that would generate semantic sequential representations (SSRs) without cue phrases.

Taboada *et al*. (2008) leverage discourse to identify relevant sentences in the text for sentiment analysis. However, they narrow their focus to adjectives alone in the relevant portions of the text while ignoring the remaining parts of speech of the text.

Most of these discourse based works make use of a *discourse parser* or a *dependency parser* to identify the scope of the discourse relations and the opinion frames. As said before, the parsers fare poorly in the presence of noisy text like *ungrammatical sentences* and *spelling mistakes* (Dey *et al.*, 2009). In addition, the use of parsing slows down any real-time interactive system due to increased processing time. For this reason, the micro-blog applications mostly build on a bag-of-words model.

## 1.9 FEATURE SPECIFIC SENTIMENT ANALYSIS

Chen *et al.* (2010) uses dependency parsing and shallow semantic analysis for Chinese opinion related expression extraction. They categorize relations as, Topic and sentiment located in the same sub-sentence and quite close to each other (like rule "an adjective plus a noun" is mostly a potential opinion-element relation), Topic and sentiment located in adjacent sub-sentences and the two sub-sentences are parallel in structure (that is to say, the two adjacent sub-sentences are connected by some coherent word, like although/but, and *etc*), Topic and sentiment located in different sub-sentences, either being adjacent or not, but the different sub sentences are independent of each other, no parallel structures any more.

Wu *et al.* (2009) use phrase dependency parsing for opinion mining. In dependency grammar, structure is determined by the relation between a head and its dependents. The dependent is a modifier or complement and the head plays a more important role in determining the behaviors of the pair. The authors want to compromise between the information loss of the word level dependency in dependency parsing as it does not explicitly provide local structures and syntactic categories of phrases and the information gain in extracting long distance relations. Hence they extend the dependency tree node with phrases."

Hu *et al.* (2004) used frequent item sets to extract the most relevant features from a domain and pruned it to obtain a subset of features. They extract the nearby adjectives to a feature as an *opinion word* regarding that feature. Using a seed set of labeled Adjectives, which they manually develop for each domain, they further expand it using WordNet and use them to classify the extracted opinion words as positive or negative.

Lakkaraju *et al.* (2011) propose a joint sentiment topic model to probabilistically model the set of features and sentiment topics using HMM-LDA. It is an unsupervised system which models the distribution of features and opinions in a review and is thus a generative model.

Most of the works mentioned above require labeled datasets for training their models for each of the domains. If there is a new domain about which no prior information is available or if there are mixed reviews from multiple domains inter-mixed (as in *Twitter*), where the domain for any specific product cannot be identified, then it would be difficult to train the models. The works do not exploit the fact that majority of the reviews have a lot of domain independent components. If those domain independent parameters are used to capture the associations between features and their associated opinion expressions, the models would capture majority of the feature specific sentiments with minimal data requirement.

## 1.10 SEMANTIC SIMILARITY METRICS

Various approaches for evaluating the similarity between two words can be broadly classified into two categories: edge-based methods and informationcontent-based methods. One of the earliest works in edge-based calculation of similarity is by Rada et al. (1989), where in, they propose a metric"Distance" over a semantic net of hierarchical relations as the shortest path length between the two nodes. This has been the basis for all the metrics involving simple edge-counting to calculate the distance between two nodes. However, the simple edge-counting fails to consider the variable density of nodes across the taxonomy. It also fails to include relationships other than the is-a relationship, thus, missing out on important information in a generic semantic ontology, like WordNet.

In contrast to edge-based methods, Richardson et al. (1994) and Resnik (1995*a*) propose a node-based approach to find the semantic similarity. They approximate conceptual similarity between two WordNet concepts as the maximum information content among classes that subsume both the concepts. Resnik (1995*b*) advanced this idea by defining the information content of a concept based on the probability of encountering an instance of that concept. Alternatively, Wu & Palmer (1994) compare two concepts based on the length of the path between the root of the hierarchy and the least common subsumer of the concepts.

Jiang & Conrath (1997) and Leacock et al. (1998) combine the above two approaches by using the information content as weights for the edges between concepts. They further reinforce the definition of information content of a concept by adding corpus statistical information.

Instead of measuring the similarity of concepts, some other approaches measure their relatedness. Hirst & St-Onge (1997) introduce an additional notion of direction along with the length of paths for measuring the relatedness of two concepts. Banerjee & Pedersen (2003) and Patwardhan (2003) leverage the gloss information present in WordNet in order to calculate the relatedness of two concepts. Banerjee & Pedersen (2003) assigns relatedness scores based on the overlap between the gloss of the two concepts. Patwardhan (2003) use a vector representation of the gloss, based on the context vector of the terms in the gloss. The relatedness is then the cosine between the gloss vectors of the two concepts.

Our work is most related to the work of Wan & Angryk (2007) which improves on Banerjee & Pedersen (2003) and Patwardhan (2003) by including relations other than the is-a relationship. They use an extended gloss definition for a concept which is defined as the original gloss appended by the gloss of all the concepts related to the given concept. They create concept vectors for each sense based on which they create context vectors which are an order higher to the concept vectors. Finally, they use cosine of the angle between the vectors of the different concepts to find their relatedness. This approach is better than other approaches as it captures the context of the concepts to a much larger extent. However, all these methods lack on a common ground. They fail to incorporate sentiment information in calculating the similarity/relatedness of two concepts. We postulate that sentiment information is crucial in finding the similarity between two concepts.

## 1.11 SENTIMENT ANALYSIS IN TWITTER

Twitter is a micro-blogging website and ranks second amongst the present social media websites (Prelovac 2010). A micro-blog allows users to exchange small elements of content such as short sentences, individual pages, or video links. Alec *et al.* (2009) provide one of the first studies on sentiment analysis on micro-blogging websites. Barbosa *et al.* (2010) and Bermingham *et al.* (2010) both cite noisy data as one of the biggest hurdles in analyzing text in such media. Alec *et al.* (2009) describes a distant supervision-based approach for sentiment classification. They use hashtags in tweets to create training data and implement a multi-class classifier with topic-dependent clusters. Barbosa *et al.* (2010) propose an approach to sentiment analysis in Twitter using POS-tagged n-gram features and some Twitter specific features like hashtags. Our system is inspired from *C-Feel-IT*, a Twitter based sentiment analysis system (Joshi *et al.*, 2011). However, our system is an enhanced version of their rule based system with specialized modules to tackle Twitter spam, text normalization and entity specific sentiment analysis.

To the best of our knowledge, there has not been any specific work regarding spam filtering for tweets in the context of sentiment analysis. General spam filtering techniques include approaches that implement Bayesian filter (Sahami, 1998; Graham, 2006), or SVM-based filters along with various boosting algorithms to further enhance the accuracies (Drucker *et al.,* 1999).

Twitter is a very noisy medium. However, not much work has been done in the area of text normalization in the social media especially pertaining to Twitter. But there has been some work in the related area of SMS-es. Aw *et al.,* (2006) and Raghunathan *et al.,* (2009) used a MT-based system for text normalization. Choudhury *et al.,* (2007) deployed a HMM for word-level decoding in SMS-es; while Catherine *et al.,* (2008) implemented a combination of both by using two normalization systems: first a SMT model, and then a second model for speech recognition system. Another approach to text normalization has been to consider each word as a corrupt word after being passed through a noisy channel, which essentially boils down to spell-checking itself. Mayes (1991) provide one such approach. Church *et al.,* (1991) provide a more sophisticated approach by associating weights to the probable edits required to correct the word.

We follow the approach of Church *et al.,* (1991) and attempt to infuse linguistic rules within the minimum edit distance (Levenstein, 1966). We adopt this simpler approach due to the lack of publicly available parallel corpora for text normalization in Twitter.

Unlike in Twitter, there has been quite a few works on general entity specific sentiment analysis. Nasukawa *et al.,* (2003) developed a lexicon and sentiment transfer rules to extract sentiment phrases. Mullen *et al.,* (2004) used Osgood and Turney values to extract value phrases, *i.e.* sentiment bearing phrases from the sentence. Many approaches have also tried to leverage dependency parsing in entity-specific SA. Mosha (2010) uses dependency parsing and shallow semantic analysis for Chinese opinion related expression extraction. Wu *et al.,* (2009) used phrase dependency parsing for opinion mining. Mukherjee *et al.* (2012) exploit dependency parsing for graph based clustering of opinion expressions about various features to extract the opinion expression about a target feature. We follow a dependency parsing based approach for entity specific SA as it captures long distance relations, syntactic discontinuity and variable word order, as is prevalent in Twitter.

The works (Alec *et al.*, 2009; Read *et al.*, 2005; Pak *et al.*, 2010; Gonzalez *et al.* (2011)) evaluate their system on a dataset crawled and auto-annotated based on *emoticons*, *hashtags*. We show, in this work, that a good performance on such a dataset does not ensure a similar performance in a general setting.

## 1.12 EXTRACTIVE SUMMARIZATION

There are 2 prominent paradigms in automatic text summarization (Das *et al.*, 2007): *extractive* and *abstractive* text summarization. While *extractive* text summarization attempts to identify prominent sections of a text by giving more emphasis on the content of the summary, *abstractive* text summarization gives more emphasis on the form so that sentences are syntactically and semantically coherent. The *topic-driven* summarization paradigm is more common to IR where the summary content is based on the user query about a particular topic. Luhn (1958) attempts to find the top-ranked significant sentences based on the frequency of the content words present in it. Edmundson (1969) gives importance to the position of a sentence *i.e.* where the sentence appears in the text and comes up with an optimum position policy and emphasis on the cue words. Aone *et al.* (1999) use tf-idf to retrieve signature words, NER to retrieve tokens, shallow discourse analysis for cohesion and also use synonym and morphological variants of lexical terms using WordNet. Lin (1999) uses a rich set of features for the creation of feature vector like *Title, Tf & Tf-Idf scores, Position score, Query Signature, IR Signature, Sentence Length, Average Lexical Connectivity, Numerical Data, Proper Name, Pronoun & Adjective, Weekday & Month, Quotation, First Sentence etc*. and use decision tree to learn the feature weights. There are other works based on HMM (Conroy *et al.*, 2008), RTS (Marcu, 1998), lexical chain and cohesion (Barzilay *et al.*, 1997).

## 1.13 SUBJECTIVITY ANALYSIS

Yu *et al.* (2003) propose to find subjective sentences using lexical resources where the authors hypothesize that subjective sentences will be more similar to opinion sentences than to factual sentences. As a measure of similarity between two sentences they used different measures including shared words, phrases and the WordNet. Potthast *et al.* (2010) focus on extracting top sentiment keywords which is based on Pointwise Mutual Information (PMI) measure (Turney, 2002).

The pioneering work for subjectivity detection is done in (Pang *et al.*, 2004), where the authors use min-cut to leverage the *coherency* between the sentences. The fundamental assumption is that local proximity preserves the objectivity or subjectivity relation in the review. But the work is completely supervised requiring two levels of tagging. Firstly, there is tagging at the sentence level to train the classifier about the subjectivity or objectivity of individual sentences. Secondly, there is tagging at the document level to train another classifier to distinguish between positive and negative reviews. Hence, this requires a lot of manual effort. Alekh *et al.* (2005) integrate graph-cut with linguistic knowledge in the form of WordNet to exploit similarity in the set of documents to be classified.

## 1.14 CONCEPT EXPANSION USING WIKIPEDIA

Wikipedia is used in a number of works for concept expansion in IR, for expanding the query signature (Muller *et al.*, 2009; Wu *et al.*, 2008; Milne *et al.*, 2007) as well as topic driven multi document summarization (Wang *et al.*, 2010).

There has been a few works in sentiment analysis using Wikipedia (Gabrilovich *et al.*, 2006; Wang *et al.*, 2008). Gabrilovich *et al.* (2006) focus on concept expansion using Wikipedia where they expand the feature vector constructed from a movie review with related concepts from the Wikipedia. This increases accuracy as it helps in unknown concept classification due to expansion but it *does not* address the concern of separating subjective concepts from objective ones.

These works do not take advantage of the Ontological and Sectional arrangement of the Wikipedia articles into categories. Each Wikipedia movie article has sections like *Plot*, *Cast, Production etc.* which can be explicitly used to train a system about the different aspects of a movie. In this work, our objective is to develop a system that classifies the *opinionated extractive summary* of the movie, requiring no labeled data for training; where the summary is created based on the extracted information from Wikipedia.

## 1.15 CONCLUSIONS

This report discusses in details the various approaches to Sentiment Analysis, mainly Machine Learning and Cognitive approaches. It provides a detailed view of the different applications and potential challenges of Sentiment Analysis that makes it a difficult task.

We have seen the applications of machine learning techniques like Naïve Bayes, Maximum Entropy, Support Vector Machines and Voted Perceptrons in SA and their potential drawbacks. As all of these are bag-of-words model, they do not capture context and do not analyze the discourse which is absolutely essential for SA. We have also seen the use of Subsequence Kernels in Voted Perceptrons that is somewhat successful to capture context as a result of which it achieves a high accuracy. Also it achieves the difficult task of performing prediction over a continuum even though trained only on the extreme reviews. Thus machine learning models with a proper kernel that can capture the context will play an important role in SA.

Feature engineering, as in several Machine Learning and Natural Language Processing applications, plays a vital role in SA. We have seen the use of phrases as well as words as features. It has been seen that Adjectives as word features can capture majority of the sentiment. Use of topic-oriented features and Value Phrases play a significant role to detect sentiment when the domain of application is known. It is also seen that use of lemmas capture sentiment better than using unigrams.

We have also discussed in details the application of Cognitive Psychology in SA. The reason why it is absolutely essential for SA is for its power of analyzing the discourse. Discourse analysis, as we have seen, plays a significant role in detecting sentiments. The use of discourse analysis and tracking point of view are necessary for analyzing opinions in blogs, newspaper and articles where a third person narrates his/her views.

We also discus some specific topics in Sentiment Analysis and the contemporary works in those areas.